\documentclass[letterpaper]{article} 
\usepackage{aaai2026}  
\usepackage{times}  
\usepackage{helvet}  
\usepackage{courier}  
\usepackage[hyphens]{url}  
\usepackage{graphicx} 
\usepackage{natbib}  
\usepackage{caption} 
\usepackage{comment}
\usepackage{booktabs}
\usepackage{subcaption} 
\begin{document}
\title{Privacy-Preserving Computer Vision for Industry:\\ Three Case Studies in Human-Centric Manufacturing}
\author{
    Sander De Coninck\textsuperscript{\rm 1},
    Emilio Gamba\textsuperscript{\rm 2},
    Bart Van Doninck\textsuperscript{\rm 2}, \\
    Abdellatif Bey-Temsamani\textsuperscript{\rm 2},
    Sam Leroux\textsuperscript{\rm 1},
    Pieter Simoens\textsuperscript{\rm 1}
}
\affiliations{
    \textsuperscript{\rm 1}IDLab, Department of Information Technology at Ghent University – imec\\


    Technologiepark 126, B-9052 Ghent, Belgium\\
    first.lastname@ugent.be

    \textsuperscript{\rm 2}Flanders Make, corelab ProductionS, Oude Diestersebaan 133, 3920 Lommel, Belgium \\
    first.lastname@flandersmake.be

%
}
\maketitle

\begin{abstract}
The adoption of AI-powered computer vision in industry is often constrained by the need to balance operational utility with worker privacy. Building on our previously proposed privacy-preserving framework, this paper presents its first comprehensive validation on real-world data collected directly by industrial partners in active production environments. We evaluate the framework across three representative use cases: woodworking production monitoring, human-aware AGV navigation, and multi-camera ergonomic risk assessment. The approach employs learned visual transformations that obscure sensitive or task-irrelevant information while retaining features essential for task performance. Through both quantitative evaluation of the privacy–utility trade-off and qualitative feedback from industrial partners, we assess the framework’s effectiveness, deployment feasibility, and trust implications. Results demonstrate that task-specific obfuscation enables effective monitoring with reduced privacy risks, establishing the framework’s readiness for real-world adoption and providing cross-domain recommendations for responsible, human-centric AI deployment in industry.

\end{abstract}

\section{Introduction}

The adoption of AI-powered computer vision in industrial settings is a cornerstone of the Industry~4.0 transformation~\cite{9667102} and a key enabler of the human-centric vision of Industry~5.0~\cite{su16135448}. From real-time ergonomic assessments to ensuring safe interactions between humans and automatic guided vehicles (AGVs), these systems promise significant gains in efficiency and workplace safety. However, their deployment is often slowed or halted altogether by a persistent sociotechnical barrier: the tension between data-driven monitoring and worker privacy~\cite{app12136395}. Addressing this challenge requires privacy to be embedded into system design from the outset. Privacy by Design (PbD)~\cite{cavoukian2009privacy} treats privacy not as a reactive safeguard but as a proactive and default principle, which is essential for creating trustworthy and human-aligned autonomous systems~\cite{LI2025308}.

Existing privacy-preserving methods rarely satisfy the PbD principle of data minimization, which requires that only task-relevant information be collected and processed. Simple obfuscation techniques such as blurring or pixelation indiscriminately degrade visual fidelity and compromise utility. More advanced approaches, including federated learning~\cite{10185053} and differential privacy~\cite{9348921}, offer theoretical guarantees but are difficult to integrate into real-time vision pipelines and do not prevent over-collection of sensitive information at the sensor level.

Our earlier work introduced a privacy-preserving computer vision framework that learns to obscure sensitive and task-irrelevant visual information through adaptive transformations~\cite{de2024privacy}. Initially validated on a pedestrian detection task, the approach demonstrated that task-specific visual filtering can balance privacy and performance. A subsequent study applied this framework to multi-camera ergonomic assessment~\cite{deconinck2025enablingprivacyawareaibasedergonomic}, extending its application to human monitoring scenarios in a more challenging setting.

To bridge the gap between theoretical framework and practical application, this paper presents the first empirical validation of our approach in real industrial settings. Specifically, we apply the framework to two new domains, woodworking production monitoring and human-aware AGV navigation, developed in collaboration with manufacturing partners using real operational data. We also revisit the ergonomics use case, incorporating feedback from an ergonomics company to evaluate practical deployment considerations. Together, these studies assess the robustness of our task-specific privacy mechanism under authentic industrial conditions.

\begin{figure}[t]
    \centering
    \includegraphics[width=\linewidth]{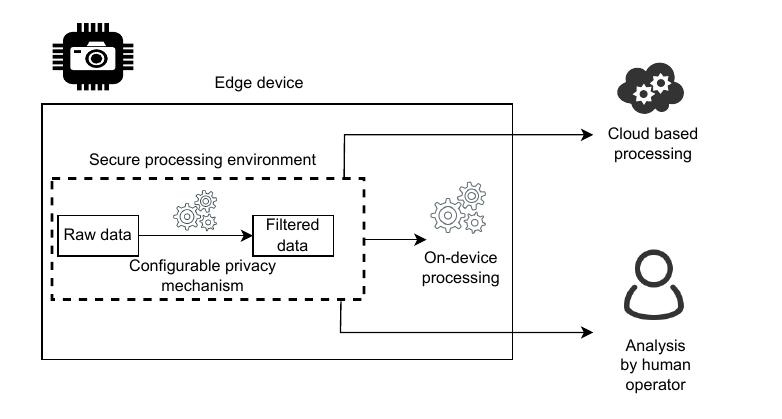}
    \caption{Conceptual overview of the proposed task-centric, privacy-preserving vision framework. 
    Raw sensor data are processed within a secure edge environment through a configurable privacy mechanism that filters task-relevant visual features before on-device or cloud-based analysis. 
    This design embeds data minimization into the sensing process, aligning with human-centric privacy principles central to Industry~5.0.}
    \label{fig:architecture}
\end{figure}

An overview of the proposed architecture is shown in Figure~\ref{fig:architecture}. The framework operates directly at the edge, where raw sensor data are processed within a secure environment using a configurable privacy mechanism that preserves only task-relevant features prior to analysis. Embedding this filtering at the sensing stage aligns with data minimization principles and reduces privacy risks, supporting responsible and human-centric deployment of AI-based computer vision in manufacturing.
\\ \newline
\noindent\textbf{Our main contributions are:}
\begin{itemize}
\item Extension and empirical evaluation of a task-centric, privacy-preserving vision framework across three industrial scenarios, including two new real-world use cases and datasets provided by industry partners;
\item Quantitative analysis of the privacy–utility trade-off using task-specific metrics;
\item Qualitative assessment of deployment readiness based on structured partner feedback;
\item Cross-domain recommendations for responsible and human-centric AI deployment in manufacturing environments.
\end{itemize}

\section{Related Work}
\label{sec:related}

\subsection{Approaches to Privacy-Preserving Visual Analysis}
Several commercial systems for industrial monitoring include simple person-blurring functions as privacy features~\cite{kamerai_privacy_2025,surveily_privacy_2025,alwaysai_privacy_2025}. However, some issues are present with this approach. For one, blurring is a naive approach that does not take into account the task at hand. This means that often important information for the task is lost, leading to a significant drop in utility~\cite{de2024privacy}. Moreover, blurring is not a strong privacy guarantee, as advanced reconstruction methods can often recover the original image~\cite{zhang2020deblurring}. Lastly, blurring persons requires accurate person detection, as one missed person leads to a complete privacy leak. 

More advanced techniques focus on inpainting or replacement. These methods typically detect sensitive entities (e.g., faces, bodies) and substitute these regions with synthetically generated content, often produced by Generative Adversarial Networks (GANs) \citep{shetty2018adversarial, hukkelas23DP2, Uittenbogaard_2019_CVPR} or diffusion models \citep{10.1145/3689943.3695048, patwari2024perceptanon}. The goal is usually to create visually plausible outputs that preserve the overall scene context while removing identifiable features. These models are often computationally more intensive, additionally, while they are able to remove person data, any contextual information (such as body shape, height, clothing) and other non-person sensitive information (such as screens, documents) is often still present in the image.

\section{A Task-Centric Adversarial Framework for Privacy Preservation}
\label{sec:framework}

To evaluate the privacy-utility trade-off in industrial settings, we apply and validate a privacy-preserving framework from our earlier work, originally developed for pedestrian detection~\cite{de2024privacy}.

\subsection{The Adversarial Obfuscation Method}
\label{subsec:method}

\begin{figure}[]
    \centering
    \includegraphics[width=\columnwidth]{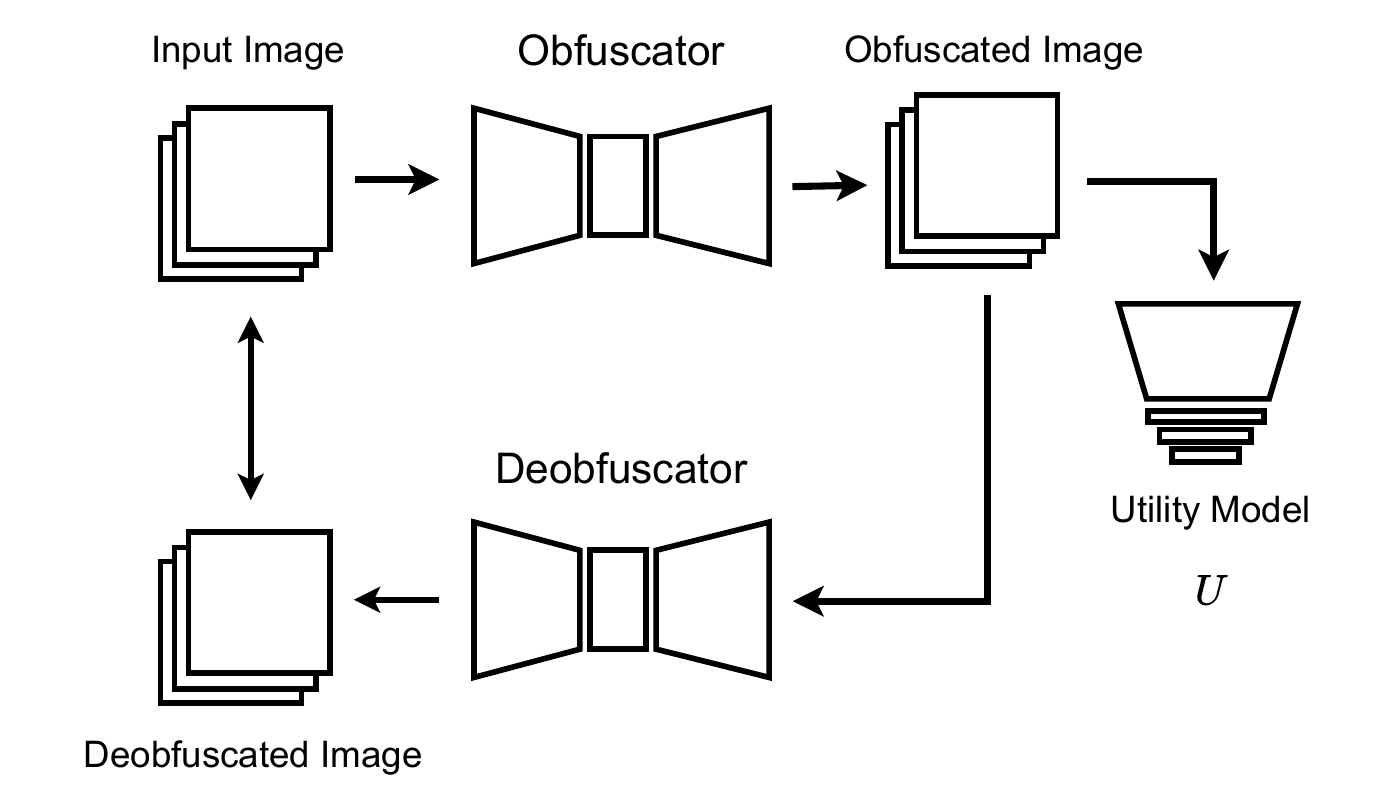}
    \caption{Overview of the adversarial framework. The Obfuscator ($O$) transforms the input image to preserve utility for the task model ($U$) while limiting the Deobfuscator’s ($D$) ability to reconstruct the original frame.}
    \label{fig:framework_diagram}
\end{figure}

The framework employs an end-to-end training process that learns a task-specific visual transformation balancing two objectives: maintaining performance on a primary ``utility'' task while protecting sensitive information from recovery. This is achieved through an adversarial setup involving three components, illustrated in Figure~\ref{fig:framework_diagram}.

\begin{enumerate}
    \item \textbf{Obfuscator (O):} A lightweight neural network that transforms a raw input frame $X$ into an obfuscated version $X'$. The goal is to sufficiently modify the image to conceal sensitive information while preserving the visual cues required for the downstream task.

    \item \textbf{Utility Model (U):} A pre-trained, off-the-shelf model that performs the target industrial task (e.g., posture classification or object detection). During training, it provides a utility-based loss signal, while its parameters remain frozen.

    \item \textbf{Adversary or Deobfuscator (D):} A reconstruction network trained to recover the original frame $X$ from its obfuscated counterpart $X'$. The adversary’s reconstruction accuracy serves as an empirical proxy for privacy risk. The more information present in the original frame, the easier it will be for the deobfuscator to reconstruct the original.
\end{enumerate}

Training proceeds by optimizing the Obfuscator $(O)$ in opposition to the Adversary $(D)$. The Obfuscator minimizes the loss associated with the Utility Model, ensuring task performance, while simultaneously maximizing the reconstruction loss of the Adversary, thereby increasing privacy protection. This adversarial objective results in a transformation that is explicitly tailored to the requirements of each industrial use case. Details of the loss functions and optimization procedure are described in~\cite{de2024privacy}\footnote{Code available at \url{https://github.com/decide-ugent/privacy-aware-ergo}}.

\subsection{Key Advantages for Industrial Contexts}
\label{subsec:rationale}

This framework is particularly well-suited for this multi-case study for five critical reasons that align directly with the challenges of real-world industrial deployment:

\begin{enumerate}
    \item \textbf{No Requirement for Sensitive Labels:} Many privacy-preserving approaches rely on explicitly labeled sensitive attributes (e.g., ``identity" or ``gender") to learn what to obscure. In contrast, this framework requires only labels for the primary utility task, such as operator keypoints or plank locations. This makes it suitable for industrial settings where datasets with fine-grained sensitive labels are rarely available and costly to obtain.

    \item \textbf{Compatibility with Existing Models:} The framework is designed to operate with frozen, pre-trained utility models. As a result, industries can retain their validated AI systems while training the obfuscator as an additional privacy layer. This modular design lowers integration effort and facilitates practical adoption.

    \item \textbf{Full-Frame Anonymization:} The Obfuscator transforms the entire input frame, not just detected individuals. This holistic approach ensures that both personal identifiers (e.g., body shape, clothing) and sensitive intellectual property (e.g., machinery, process details) are obscured, a critical feature for industrial environments.

    \item \textbf{Efficiency for Edge Deployment:} The Obfuscator itself is designed to be a lightweight model (e.g., based on MobileNet~\cite{sandler2018mobilenetv2}, as in \cite{de2024privacy}). This makes it computationally feasible to deploy on edge devices like smart cameras directly on the factory floor, processing video at the source before any data is sent to a central server.
    
    \item \textbf{Alignment with Data Minimization Principles:} The framework is designed to transform the input data to preserve only the minimal information required for the utility task, which aligns with the data minimization principle in privacy regulations like GDPR. This makes it a suitable choice for industrial applications that must comply with such regulations.
\end{enumerate}

\section{Experiments and Evaluation}
\label{sec:experiments}

To assess the real-world viability of the proposed task-centric privacy framework, we conducted a mixed-methods evaluation across three distinct industrial scenarios. Each scenario was co-designed with an industry partner to address specific privacy-related deployment challenges and relies on a newly collected dataset tailored to the corresponding utility task.

\subsection{Industrial Use Cases and Datasets}
\label{subsec:datasets}

\begin{enumerate}
    \item \textbf{Ergonomics Monitoring:} The key challenge is to preserve worker privacy while accurately classifying ergonomic risk. The system was initially studied in \cite{deconinck2025enablingprivacyawareaibasedergonomic}, where multi-camera pose estimation was used to calculate ergonomic risk scores from 3D skeletons. In this paper, we include the ergonomics scenario as part of our broader multi-case evaluation and extend the analysis by incorporating structured feedback from an industry partner (Partner A), providing insights on real-world deployment and privacy considerations.
    \begin{itemize}
        \item \textbf{Task:} Pose estimation across multiple camera views. The goal is to accurately estimate human body keypoints which can be used to calculate an ergonomic risk score like REBA~\cite{hignett2000rapid}.
        \item \textbf{Dataset:} A dataset was collected in a controlled lab environment. 20 participants, with informed consent, performed a series of lifting tasks, while being captured by 4 cameras. The footage consists of around 87000 frames.
        \item \textbf{Utility Model (U):} Ultralytics YOLOv11~\cite{Jocher_Ultralytics_YOLO_2023} pose estimation model. 
    \end{itemize}

    \item \textbf{Activity Monitoring in Woodworking (Partner B):} Partner~A, a manufacturing software company, previously terminated promising efforts to use cameras for production monitoring due to privacy concerns. They require a system that can recognize worker activities without intrusive surveillance.
    \begin{itemize}
        \item \textbf{Task:} Detection of individual planks. A conveyor belt system transports planks, detection of the planks can be used for multiple tasks such as counting machine throughput and detecting anomalies such as machine downtime.
        \item \textbf{Dataset:} A dataset was collected in a woodworking shop. Three cameras captured a conveyor belt system transporting planks, where one employee is interacting with the conveyor belt. The dataset is modest in size, with 395 frames, each annotated with bounding boxes for each plank.
        \item \textbf{Utility Model (U):} A custom trained YOLOv11 model for plank detection.
    \end{itemize}

    \item \textbf{Human-Aware AGV Navigation (Partner C):} Partner~B develops autonomous navigation for warehouse trucks (AGVs). Their systems use cameras to detect and avoid collisions with workers, but they must address the privacy implications of constant video capture.
    \begin{itemize}
        \item \textbf{Task:} Human detection and localization.
        \item \textbf{Dataset:} A dataset was collected in an active warehouse environment where AGVs operate. The dataset includes 1800 frames. Human bounding boxes were annotated using predictions from a pre-trained YOLOv11 model and were manually reviewed for accuracy.
        \item \textbf{Utility Model (U):} A pre-trained YOLOv11 object detection model.
    \end{itemize}
\end{enumerate}

\begin{figure*}[ht]
    \centering
    \begin{minipage}{0.48\textwidth}
        \centering
        \includegraphics[width=0.95\linewidth]{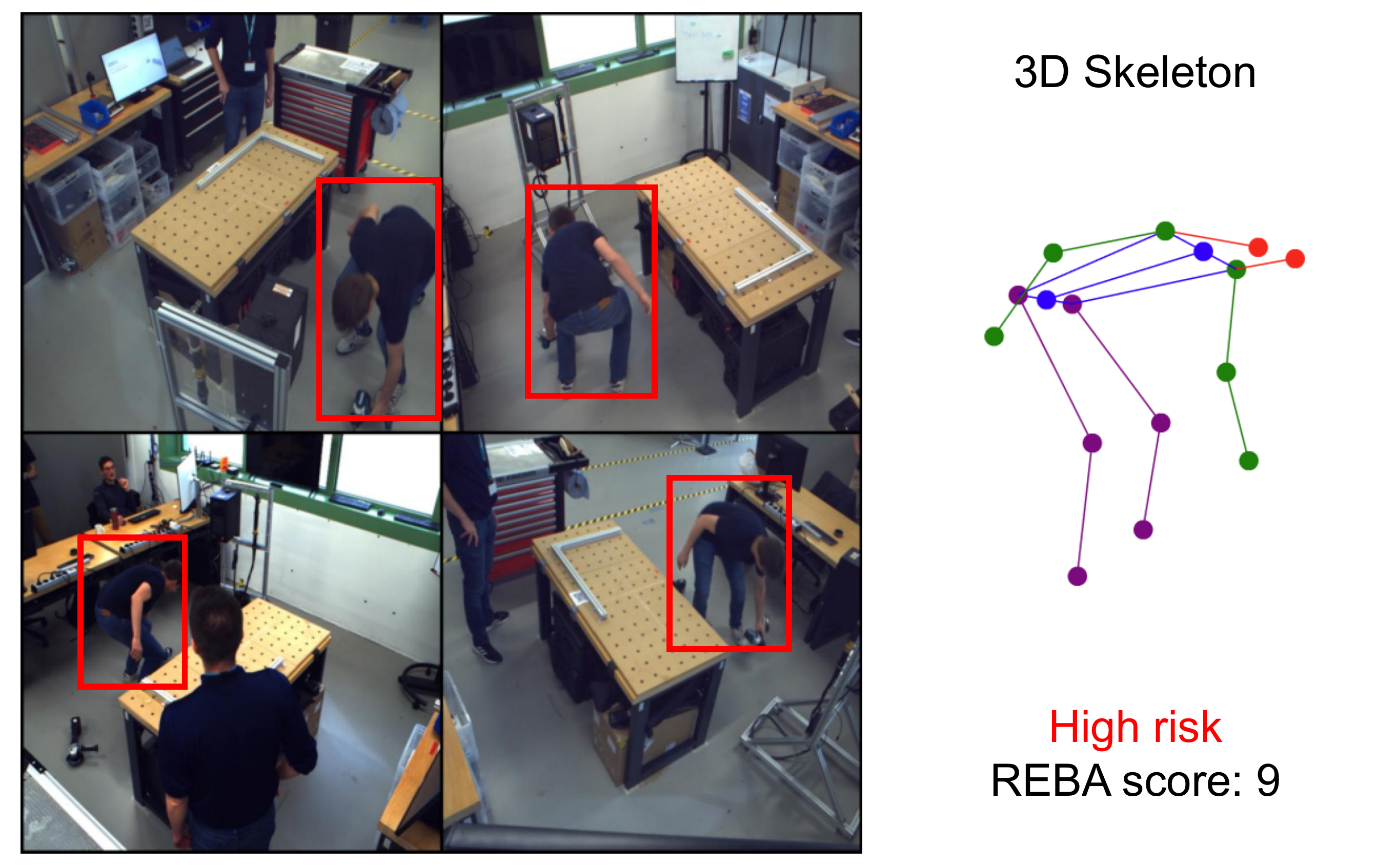} 
        {(a) Ergonomics Monitoring}
    \end{minipage}\hfill
    \begin{minipage}{0.48\textwidth}
        \centering
        \includegraphics[width=0.95\linewidth]{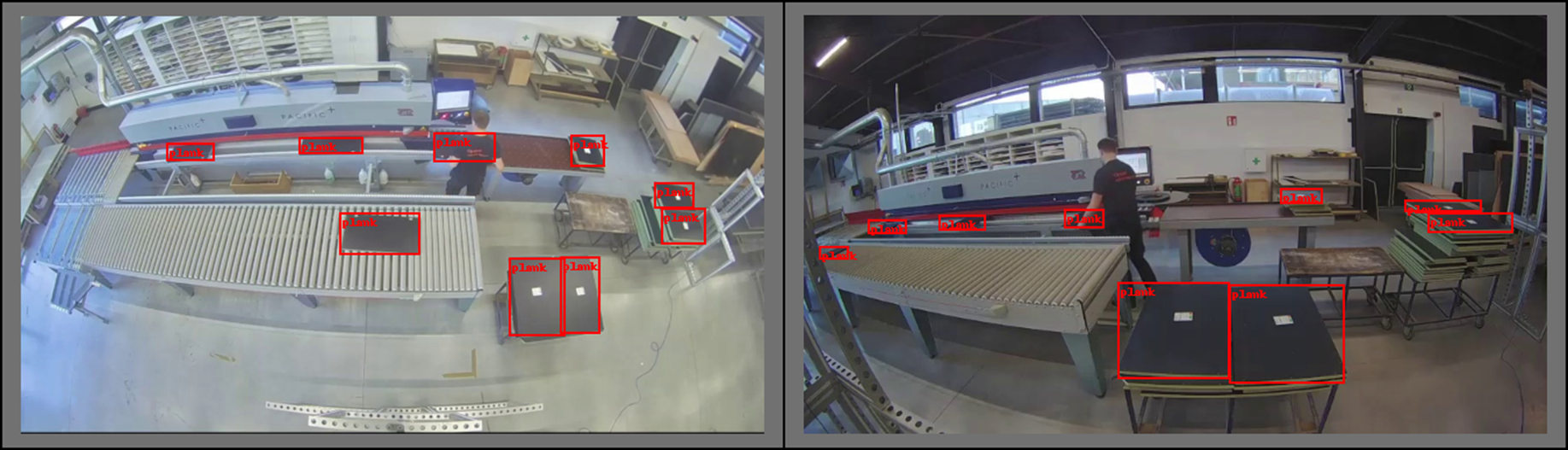} 
        {(b) Woodworking Production Monitoring}
        \vspace{2mm}
        \includegraphics[width=0.95\linewidth]{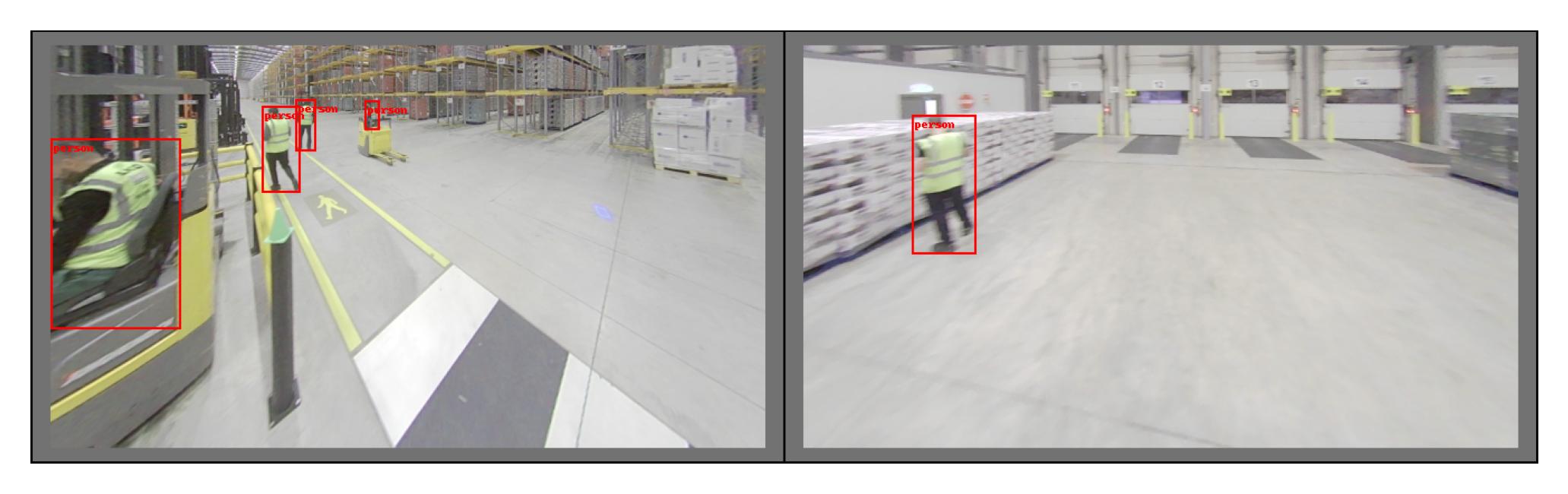} 
        { (c) AGV Pedestrian Detection}
    \end{minipage}
    \caption{Example frames from the three industrial use cases. Left: (a) Ergonomics. Right: (b) Woodworking and (c) AGV pedestrian detection.}
    \label{fig:dataset_examples}
\end{figure*}

All three cases involve monitoring industrial workspaces with cameras, raising similar privacy concerns. The ergonomics scenario revisits prior work, while the woodworking and AGV datasets were newly collected in collaboration with our industry partners. In the woodworking scenario, the task focuses on detecting planks rather than humans, but privacy concerns remain because operators interact with the planks and move through the environment. For the ergonomics and AGV scenarios, humans are the primary focus of detection, making privacy considerations even more critical. Example frames from each dataset are shown in Figure~\ref{fig:dataset_examples}. The main differences across the scenarios include the level of human involvement, camera motion, and dataset size, all of which directly impact the achievable privacy–utility trade-off, evaluated in the following sections.

\subsection{Evaluation Metrics}
\label{subsec:metrics}

Our evaluation combines quantitative metrics for both task performance and privacy preservation with qualitative feedback from industrial stakeholders to assess real-world deployment viability.

\begin{itemize}
    \item \textbf{Task Utility:} Measures the performance of the utility model on obfuscated data. The specific metric is task-dependent and reflects the practical effectiveness of the model in each industrial scenario:
\begin{itemize}
    \item \textit{Ergonomics (Pose Estimation):} Mean Average Precision (mAP) evaluated at a 0.5 Object Keypoint Similarity (OKS) threshold.
    \item \textit{Woodworking \& AGV Navigation (Object Detection):} Mean Average Precision (mAP) evaluated at a 0.5 Intersection over Union (IoU) threshold.
\end{itemize}

    \item \textbf{Privacy Preservation:} We utilize the PerceptAnon~\cite{patwari2024perceptanon} metric, which reflects human perception of anonymization by considering personal identifiers and contextual information. We report the HA2 variant, ranging from 0 to 10, where 10 indicates the highest level of privacy.
\end{itemize}

To benchmark our framework, we compare its privacy-utility trade-off against two common baselines. The first is a task-agnostic Gaussian blur applied to the entire frame with varying kernel sizes ($k$), where a larger kernel size corresponds to a stronger and more aggressive obfuscation. The second is a two-step anonymization approach where a YOLOv11 model first detects persons, and Gaussian blur is subsequently applied to the resulting bounding boxes, again across a range of kernel sizes.

\subsection{Implementation Details}
\label{subsec:implementation}

For each use case, we implemented the adversarial obfuscation framework as described. We performed an extensive hyperparameter search to optimize the Obfuscator's performance, focusing on the balance between utility and privacy. Important hyperparameters include the size of the obfuscator model, the learning rate of both obfuscator and deobfuscator, and the parameter impact of deobfuscator reconstruction performance. All training was performed on a single Tesla V100 GPU. While training incurs significantly higher computational costs, inference requires only the Obfuscator model, making real-time edge deployment feasible in industrial environments.

\subsection{Quantitative Results: The Privacy-Utility Trade-off}
\label{subsec:quant_results}

\begin{figure*}[t]
    
    \centering
    \includegraphics[width=2\columnwidth]{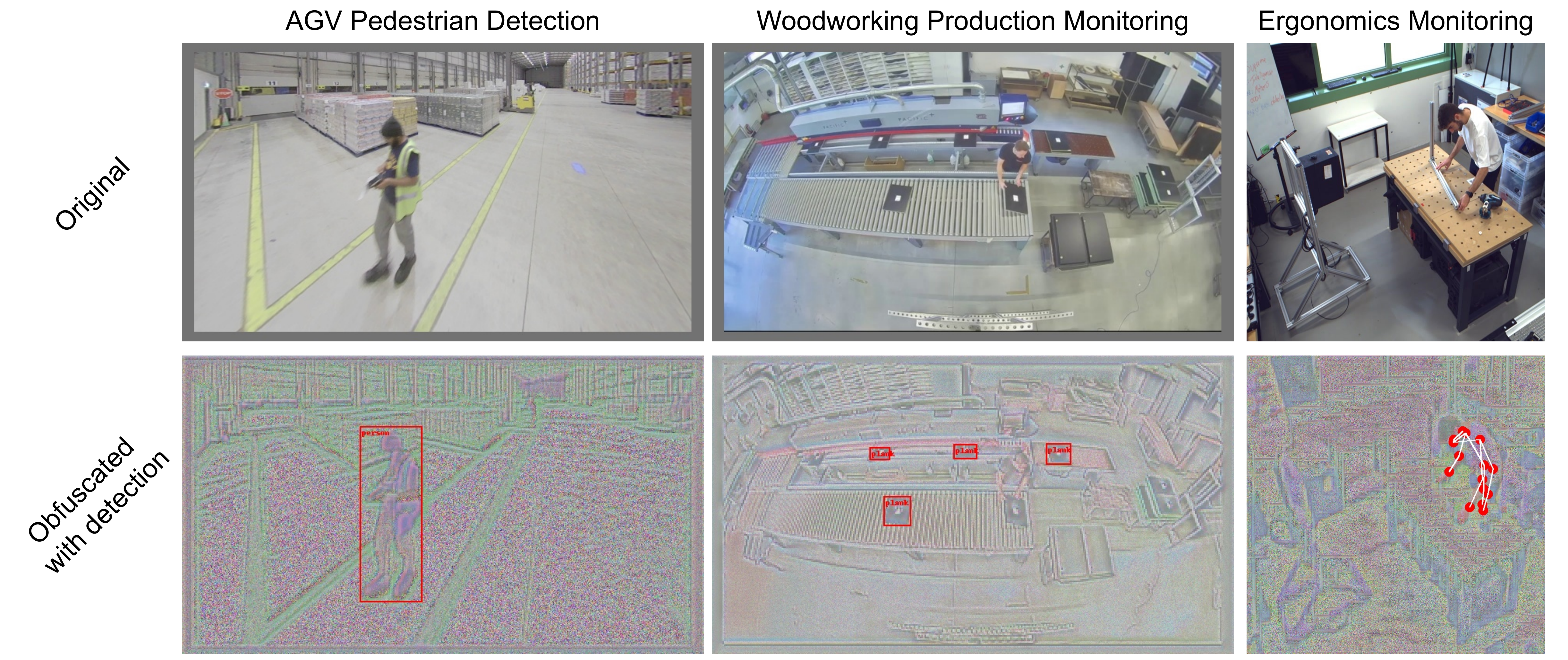}
    \caption{Original and obfuscated with detection examples for each use case. Note that detections are made by the original models that have not been retrained on obfuscated data.}
    \label{fig:example_figs}
\end{figure*}

\begin{figure*}[t!]
    \centering
    \begin{subfigure}[b]{0.32\textwidth}
        \centering
        \includegraphics[width=\textwidth]{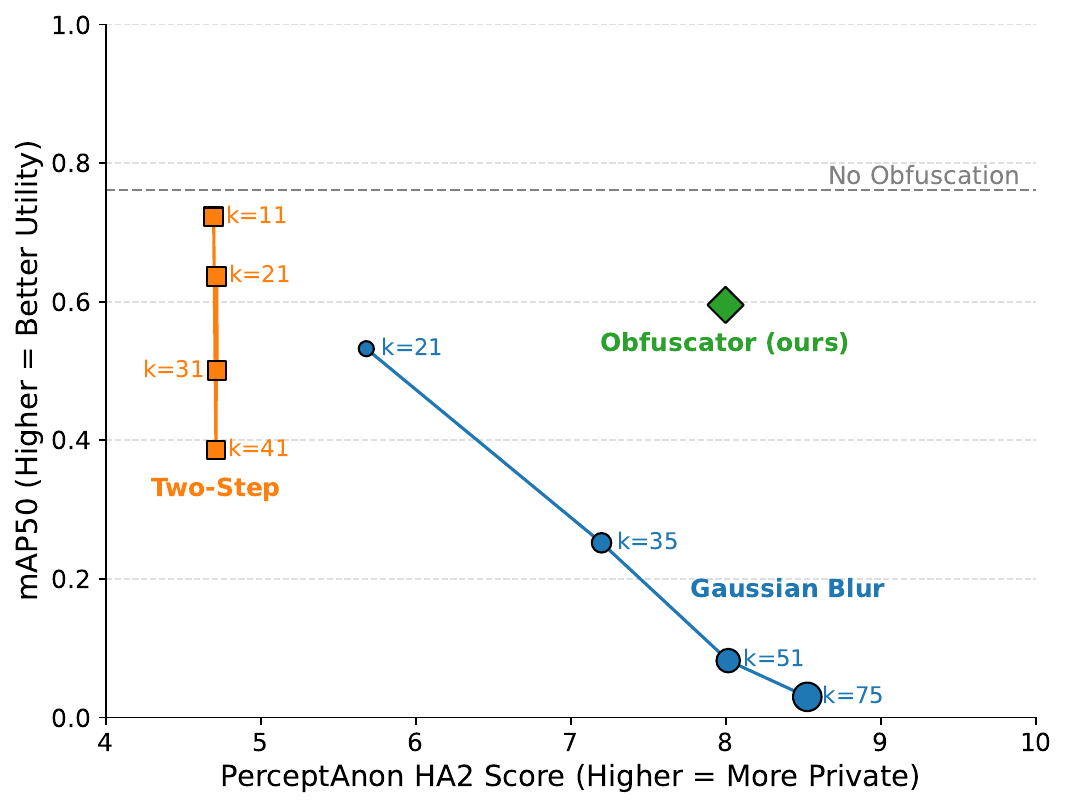}
        \caption{AGV Pedestrian Detection.}
        \label{fig:agv}
    \end{subfigure}
    \hfill 
    \begin{subfigure}[b]{0.32\textwidth}
        \centering
        \includegraphics[width=\textwidth]{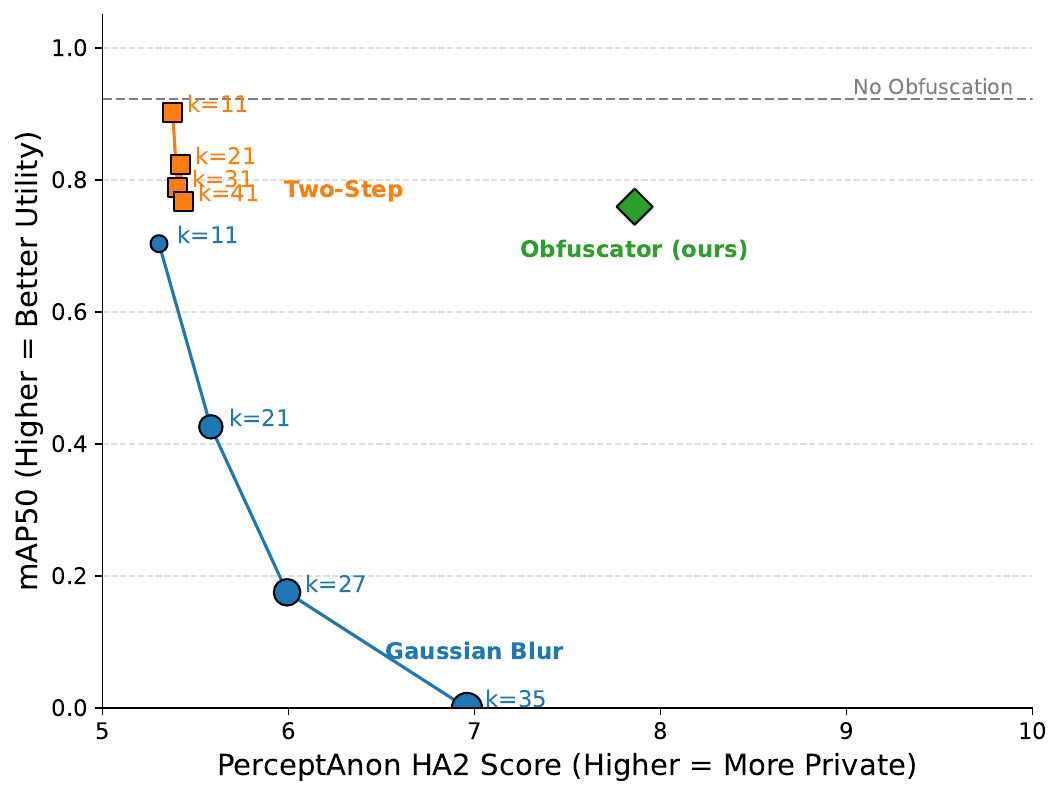}
        \caption{Woodworking Production Monitoring.}
        \label{fig:wood}
    \end{subfigure}
    \hfill 
    \begin{subfigure}[b]{0.32\textwidth}
        \centering
        \includegraphics[width=\textwidth]{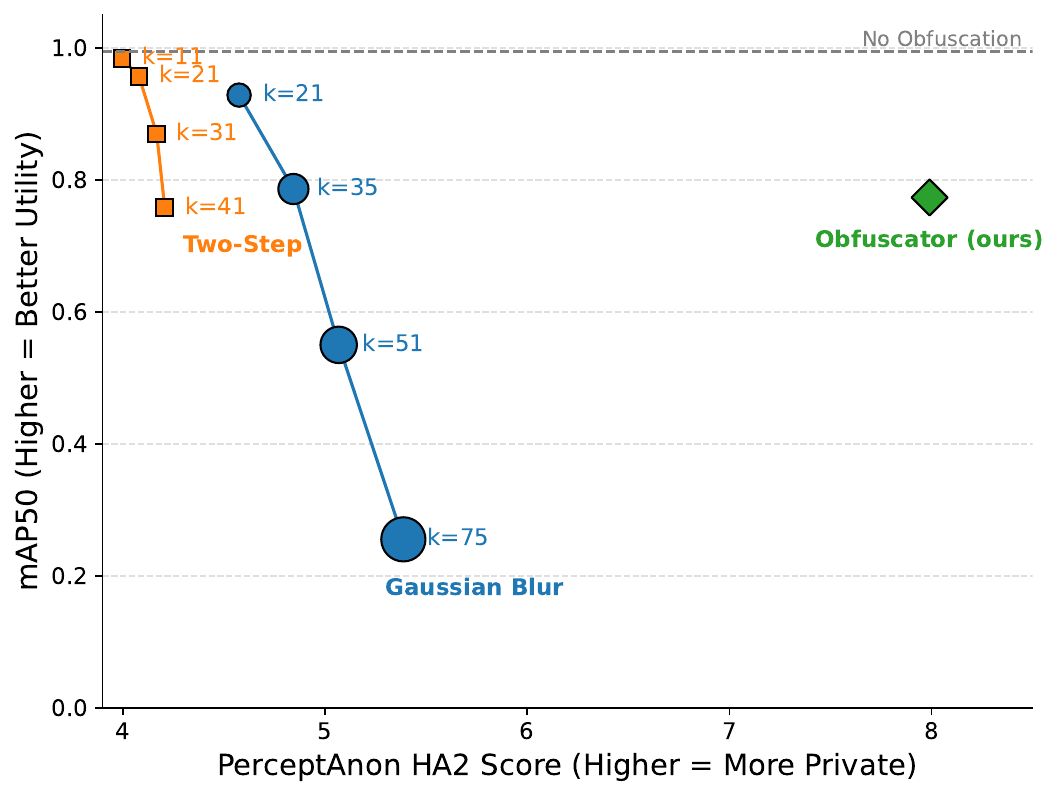}
        \caption{Ergonomics Monitoring }
        \label{fig:ergo}
    \end{subfigure}
    
    \caption{Privacy-utility curves for the three use-cases. We compare our obfuscator against basic techniques such as whole-image blurring and a two-step technique where persons are first detected and then blurred. Our technique achieves a more optimal privacy-utility tradeoff than these basic techniques.}
    \label{fig:combined_results}
\end{figure*}

Examples of the visual output from our framework are shown in Figure~\ref{fig:example_figs}. 
The quantitative evaluation of its performance is detailed in the privacy-utility curves of Figure~\ref{fig:combined_results}, which compare our task-centric obfuscator against two common baselines: task-agnostic Gaussian blur and a two-step, detect-then-blur technique.

The results clearly demonstrate that our method achieves a superior privacy-utility trade-off across all three industrial scenarios. In each case, our method (green diamond) is positioned in the desirable top-right region of the plot, achieving a significantly higher utility (mAP) for a given privacy level (PerceptAnon score) than either baseline. For instance, in the AGV navigation scenario (Figure~\ref{fig:agv}), our method delivers an mAP score of approximately 0.6 at a privacy level of 8.0. In stark contrast, the Gaussian blur method fails to exceed an mAP of 0.2 at a comparable privacy level, whereas the two-step method fails to even reach that level of privacy protection, confirming the significant value of a learned, task-specific transformation.

While our method consistently outperforms the baselines, its absolute effectiveness varies across the use cases. The ergonomics scenario demonstrates the most favorable trade-off, which we attribute to the larger volume of training data and the use of static cameras. Conversely, the AGV pedestrian detection scenario exhibits a more pronounced reduction in utility. This decrease is likely due to the smaller dataset and the challenges of a moving camera, which requires the obfuscation model to generalize across more varied viewpoints and backgrounds.

\begin{figure}[t]
    \centering
    \includegraphics[width=0.9\columnwidth]{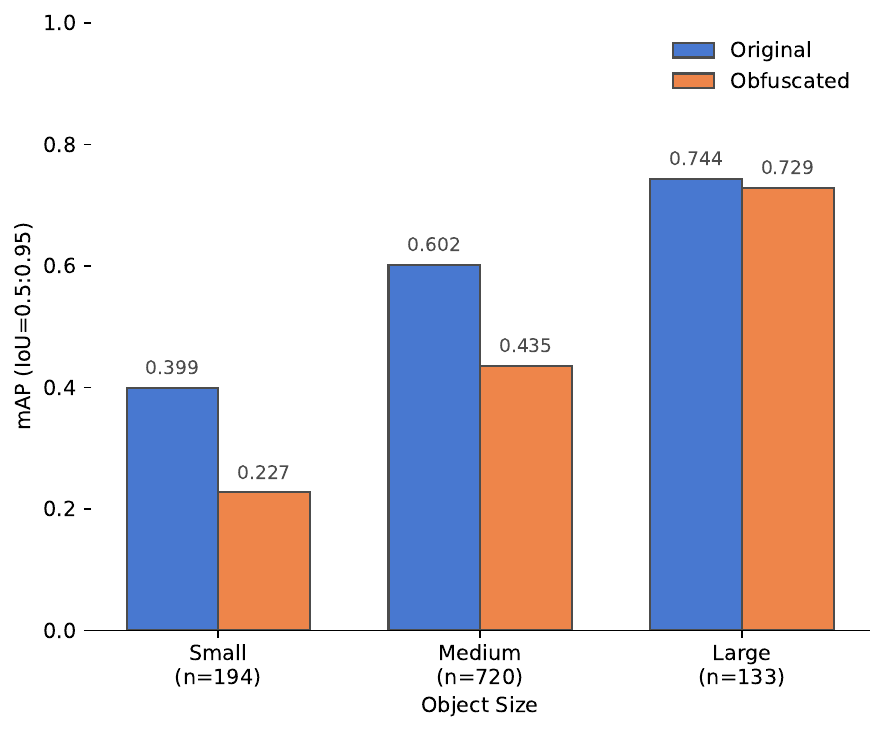}
    \caption{Woodworking: mAP vs. object size. Smaller objects suffer larger post-obfuscation drops.}
    \label{fig:size_vs_map_wood}
\end{figure}

\begin{figure}[t]
    \centering
    \includegraphics[width=0.9\columnwidth]{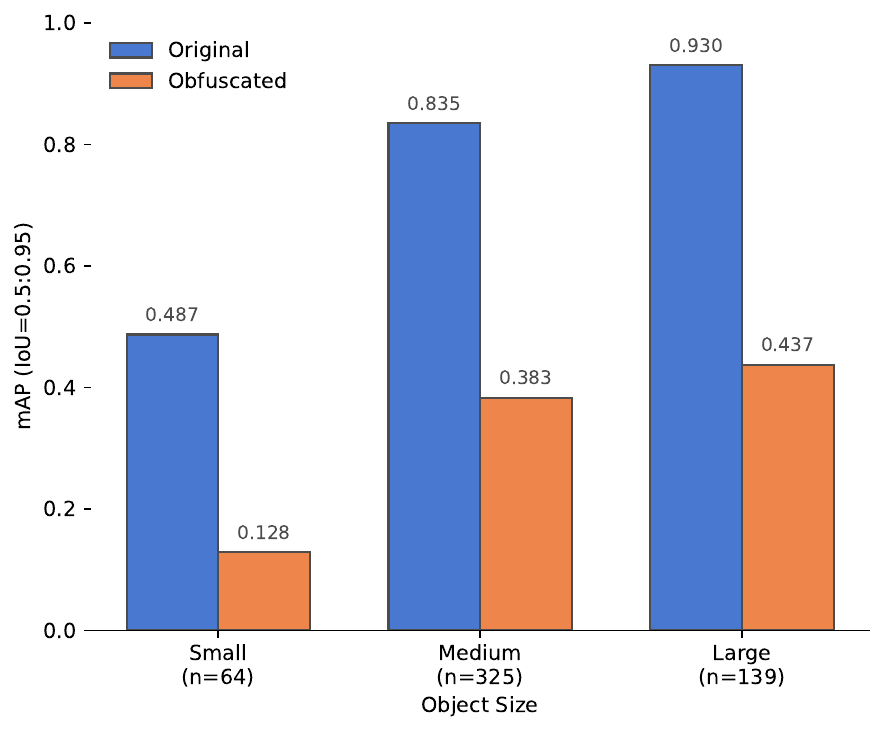}
    \caption{AGV pedestrian detection: mAP vs. object size. Smaller objects suffer larger post-obfuscation drops, although the difference between large and medium objects is less pronounced.}
    \label{fig:size_vs_map_agv}
\end{figure}

To further dissect the factors influencing utility, we conducted an additional experiment on the woodworking and AGV scenarios to study the effect of object size on performance. The results in Figures~\ref{fig:size_vs_map_wood} and~\ref{fig:size_vs_map_agv} reveal that smaller objects are disproportionately affected by obfuscation. In the woodworking case (Figure~\ref{fig:size_vs_map_wood}), the mAP for large objects drops minimally from 0.744 to 0.729 (a 2\% reduction), while the mAP for small objects plummets from 0.399 to 0.227 (a 43\% reduction). A similar, though less pronounced, trend is observed in the AGV scenario (Figure~\ref{fig:size_vs_map_agv}). This indicates that the fine-grained features necessary to detect small objects are more vulnerable to information loss during the privacy-preserving transformation, a key consideration for future improvements. We note that the baseline accuracies in the AGV case are higher, as its labels were semi-automatically generated by a pre-trained YOLOv11 model.

\subsection{Qualitative Results: Stakeholder Feedback for Deployment Viability}
\label{subsec:qual_results}

To assess the real-world viability of our framework beyond quantitative metrics, we conducted structured interviews with our industrial partners in ergonomics, woodworking, and autonomous navigation. These discussions revealed that successful adoption hinges on a delicate balance of socio-technical factors, including operational context, regulatory acceptance, and worker trust, which often prove more critical than purely algorithmic performance.

\subsubsection{Ergonomic Risk Assessment}

For the ergonomic risk assessment, Partner A highlighted a key trade-off inherent in the full-frame obfuscation approach. While the method effectively preserves privacy for the primary task of pose estimation, the obfuscation of the entire scene also obscures environmental details like tools or workstation height. This presents a challenge for more comprehensive ergonomic assessments that rely on analyzing an operator's full interaction with their surroundings.

Second, operators may feel uneasy about continuous monitoring with multiple cameras, even when privacy-preserving measures are implemented and the underlying task is to ensure well-being of the operator. Establishing and maintaining trust among operators, management, and monitoring service providers is therefore critical. Trust-building mechanisms include real-time transparency, such as displaying the privacy-aware, obfuscated video feed directly on the workstation, allowing operators to verify what data is collected and how their privacy is preserved.

Partner A also recommended implementing access controls that enable operators to review collected data and understand how it informs workplace improvements, fostering a positive feedback loop that emphasizes collaborative benefits rather than surveillance.

\subsubsection{Woodworking Process Monitoring} 
In the woodworking scenario (\textit{Partner B}), privacy was regarded as a prerequisite for reintroducing camera-based monitoring, which had previously been discontinued due to concerns raised by workers. A question emerged regarding whether simpler anonymization approaches, such as person blurring, could be adopted, given that our method entails a certain degree of accuracy loss. However, such approaches may inadvertently obscure critical information, particularly when individuals are positioned close to relevant objects or areas of interest. A more rigorous analysis is therefore required to evaluate the trade-offs between obfuscation-induced losses and those arising from person-based anonymization. Nonetheless, our obfuscation method offers more consistent privacy guarantees, as it does not depend on the accurate detection of individuals.

A key insight was that the effectiveness of privacy protection depends strongly on contextual factors. Obfuscation in environments with only a small number of workers is inherently less effective, as the likelihood of re-identification increases. Partner~B proposed the introduction of user-configurable privacy levels, enabling end-users to choose between:
\begin{enumerate}
    \item \textit{High-assurance mode:} prioritizing robust anonymization to minimize re-identification risks in sensitive contexts, even at the expense of reduced utility;
    \item \textit{Standard mode:} emphasizing operational performance while maintaining general compliance in lower-risk environments.
\end{enumerate}
This perspective reconceptualizes the privacy–utility trade-off as a dynamic control parameter, promoting more human-centric and transparent privacy management.

Partner~B noted that certain tasks in the woodworking workflow require high-resolution imagery ($>640\times640$) to ensure sufficient visual detail for operational purposes. However, such resolutions are currently not supported by our obfuscation algorithm due to computational constraints. Further research is needed to explore potential optimizations that could enable high-resolution support without compromising privacy guarantees.

\subsubsection{Autonomous Navigation}
In the autonomous navigation scenario (\textit{Partner C}), the focus shifted toward safety-critical operation. Here, traditional performance metrics such as mean Average Precision were found insufficient to capture true task utility. Stakeholders emphasized that recall, the ability to detect all nearby persons, is more crucial than precision, since missed detections could lead to collisions, whereas false positives merely slow operations. They further noted that detection range matters: reliable detection within approximately 6–8\,meters is essential for safe braking, while distant detections are less critical. Real-time performance was identified as a key constraint. To enable timely navigation responses, the obfuscation pipeline must sustain processing speeds of at least 10–15\,fps at the edge. 

Regarding privacy protection, Partner~C viewed the current obfuscation strength as sufficient for meeting regulatory requirements, provided that individuals are not identifiable from either facial or behavioral cues. They stressed that regulatory acceptance often depends on demonstrable due diligence and transparency rather than formal guarantees, underscoring the importance of clear documentation and communication with workers and auditors.

\subsubsection{Cross-Partner Insights}
Across all industrial partners, several consistent themes emerged from our evaluations. First, privacy-preserving vision systems must be configurable to accommodate varying operational and regulatory contexts. Partners highlighted that traditional task metrics alone, such as mean Average Precision, often fail to capture true operational utility, emphasizing the need for multi-level evaluation frameworks that reflect real-world objectives.

Residual privacy risks were identified as a key concern. Even when visual anonymization is applied, movement-based cues such as gait, posture, or interaction patterns may allow familiar observers to re-identify individuals. This observation underscores the limitations of purely visual metrics and suggests that factors such as team size and workplace familiarity strongly influence effective anonymization.

Trust and transparency were consistently emphasized as critical for adoption. Partners noted the difficulty of explaining technical privacy mechanisms to non-technical operators. Demonstrations using live obfuscated video feeds, coupled with clear documentation of data handling practices, were recommended to ensure stakeholders understand how privacy is preserved. Third-party validation and certification, for example by universities, labor unions, or standardization bodies, were seen as essential to establish both regulatory compliance and worker confidence.

Taken together, these insights highlight that effective privacy preservation in industrial AI requires a socio-technical approach, balancing algorithmic performance with context-sensitive deployment practices and human factors.
\section{Discussion: A Pathway Toward Deployment}
\label{sec:discussion}

Our empirical results demonstrate that task-centric adversarial obfuscation can preserve utility while protecting privacy across diverse industrial settings. However, several challenges remain before these systems can be widely deployed.  

\subsection{Defining and Measuring Privacy}
Privacy is inherently context-dependent, and universally accepted metrics are lacking. Image-based measures such as SSIM~\cite{wang_image_2004}, VIF~\cite{sheikh_image_2006}, or PerceptAnon~\cite{patwari2024perceptanon} can quantify visual distortion, but they do not fully capture whether identifiable information remains. In practice, privacy assurance depends not only on formal guarantees but also on perceived compliance and transparency, as highlighted by industrial partners. Human evaluation, adversarial re-identification~\cite{hanisch2023false}, and image-similarity metrics each have limitations, reinforcing the need for multi-level, context-aware assessment strategies.  

\subsection{Task-Level Utility and Operational Relevance}
Quantitative metrics such as mAP provide a proxy for task performance, but true utility is defined by operational objectives. For example, in the woodworking case, accurate object detection supports throughput estimation and downtime detection, which may not align perfectly with model-level accuracy. Similarly, in AGV navigation, recall within a safety-critical range is more important than precision. These examples highlight the urgent need for operationally-grounded benchmarks, where system performance is measured not by generic metrics like mAP, but by its direct impact on key operational objectives.  

\subsection{Practical Constraints for Industrial Deployment}
Industrial deployment introduces constraints related to computational efficiency, scalability, and data collection. Training the obfuscator requires labeled datasets and substantial GPU resources, while inference must meet edge-device speed requirements. High-resolution imagery, common in industrial environments, further challenges obfuscation models, as observed in our experiments with resolutions above $640\times640$. Organizational factors, including maintainability, transparency, and auditability, also influence adoption. Deployment strategies must balance privacy protection with operational feasibility and regulatory compliance.  Lastly, companies need to determine effective ways to communicate how privacy preservation is implemented, through mechanisms such as visible obfuscation or clear purpose limitations, in order to support worker acceptance and trust

\subsection{Conclusion}
Effective deployment of privacy-preserving computer vision in industrial settings requires more than strong algorithms. Success depends on integrating technical obfuscation with task-aware utility metrics, operational context, and socio-technical considerations, such as transparency, configurability, and third-party validation. This holistic approach aligns with the human-centric principles of Industry~5.0~\cite{su16135448} and establishes a pathway for responsible AI adoption in manufacturing and other industrial domains.

The potential for this privacy-aware paradigm extends to any industrial task where the subject's identity is irrelevant but their actions, or the state of objects around them, are critical. This includes broad safety compliance applications, such as monitoring for personal protective equipment~\cite{nath2020deep}, as well as fine-grained operational tasks like quality control and anomaly detection~\cite{10663422}. By decoupling analysis from identification, task-centric approaches can unlock the benefits of computer vision in even the most sensitive industrial environments.

\section{Acknowledgments}

We would like to thank our industrial partners for their extensive feedback and for delivering data for our experiments. We want to explicitly thank Nick Koopmans and Bob Grietens from PixelVision, Bern De Lille from ARDIS and Luc Thomas from ExoVibe.

Sander De Coninck receives funding from the Special Research Fund of Ghent University under grant no. BOF22/DOC/093. This research is done in the framework of the Flanders AI Research Program (\url{https://www.flandersairesearch.be/en}) financed by EWI (Economie Wetenschap \& Innovatie), and Flanders Make (\url{https://www.flandersmake.be/en}), the strategic research Centre for the Manufacturing Industry who owns the Operator 4.0/5.0 infrastructure. It is part of the Flanders AI Enrichment Project EP05 - PRIVAI.

\bibliography{references}

\end{document}